\documentclass[10pt,twocolumn,letterpaper]{article}

\usepackage[utf8]{inputenc}

\usepackage{cvpr}
\usepackage{times}
\usepackage{epsfig}
\usepackage{graphicx}
\usepackage{amsmath}
\usepackage{amssymb}
\usepackage{amsthm}
\usepackage{mathtools}
\usepackage{bm}
\usepackage{microtype}
\usepackage{placeins}
\usepackage{booktabs}
\usepackage[load=abbr]{siunitx}
\usepackage[inline]{enumitem}
\usepackage{caption}
\DeclareSIUnit\frame{fr}
\usepackage{subcaption}
\usepackage{nidanfloat}
\usepackage{chngcntr}
\usepackage{mathrsfs}
\usepackage{adjustbox}

\usepackage{floatrow}
\DeclareFloatFont{tiny}{\tiny}% "scriptsize" is defined by floatrow, "tiny" not
\setlist{topsep=0.5ex,itemsep=-0.5ex,partopsep=1ex,parsep=1ex}
\makeatletter
\renewcommand{\paragraph}{%
  \@startsection{paragraph}{4}%
  {\z@}{1.7ex \@plus 0.5ex \@minus .2ex}{-1em}%
  {\normalfont\normalsize\bfseries}%
}
\makeatother

\usepackage{amsmath,amsfonts,bm}

\def\eqref#1{equation~\ref{#1}}
\def\1{\bm{1}}

\def\vzero{{\bm{0}}}

\def\vc{{\bm{c}}}

\def\vl{{\bm{l}}}

\def\vM{{\bm{M}}}

\def\vx{{\bm{x}}}

\def\vzero{{\bm{0}}}

\def\mA{{\bm{A}}}

\def\mV{{\bm{V}}}

\DeclareMathAlphabet{\mathsfit}{\encodingdefault}{\sfdefault}{m}{sl}
\SetMathAlphabet{\mathsfit}{bold}{\encodingdefault}{\sfdefault}{bx}{n}

\usepackage[pagebackref=true,breaklinks=true,colorlinks,bookmarks=false]{hyperref}
\hypersetup{pdfauthor={Maxim Berman, Leonid Pishchulin, Ning Xu, Matthew B. Blaschko, Gérard Medioni},pdftitle={AOWS: Adaptive and optimal network width search with latency constraints},pdfkeywords={Neural architecture search, Mobilenet, TensorRT, Latency, Classification, ImageNet, Viterbi, Network width search, OWS, AOWS},pdfsubject={AOWS calibrates the width of neural network architectures using slimmable networks and the Viterbi algorithm.}}
\usepackage{hypcap}
\usepackage{cleveref}
\usepackage{authblk}

\usepackage{floatrow}
\theoremstyle{definition}
\newtheorem{problem}{Problem}[section]

\theoremstyle{plain}
\newtheorem{assumption}{Assumption}[section]
\crefname{appendix}{supplementary}{supplementaries}

\newif\ifarxiv
\ifarxiv
  \usepackage{fancyhdr}
  \let\svmaketitle\maketitle
  \def\maketitle{\svmaketitle\thispagestyle{fancy}}
\fi

\cvprfinalcopy % *** Uncomment this line for the final submission

\def\cvprPaperID{2006} % *** Enter the CVPR Paper ID here

\renewcommand{\S}{\bm{\mathcal{S}}}
\newcommand{\C}{\bm{\mathcal{C}}}

\ifcvprfinal\fi

\newif\ifarxiv
\arxivtrue

\ifarxiv
  \usepackage{fancyhdr}
  \let\svmaketitle\maketitle
  \def\maketitle{\svmaketitle\thispagestyle{fancy}}
\else
  \ifcvprfinal
  \let\svmaketitle\maketitle
  \def\maketitle{\svmaketitle\thispagestyle{empty}}
  \fi
\fi

\begin{document}

%%%%%%%%% TITLE
\title{AOWS: Adaptive and optimal network width search with latency constraints\vspace{-0.5em}}

\author[1]{Maxim Berman\thanks{Work done during an internship at Amazon.}}
\author[2]{Leonid Pishchulin}
\author[2]{Ning Xu}
\author[1]{Matthew B. Blaschko}
\author[2]{Gérard Medioni}
\affil[1]{Center for Processing Speech and Images, Department of Electrical Engineering, KU Leuven}
\affil[2]{Amazon Go}

\renewcommand\Authsep{\quad}
\renewcommand\Authands{\quad}

\maketitle

%%%%%%%%% ABSTRACT
\begin{abstract}
    Neural architecture search (NAS) approaches aim at automatically finding novel CNN architectures that fit computational constraints while maintaining a good performance on the target platform. 
    We introduce a novel efficient one-shot NAS approach to optimally search for channel numbers, given latency constraints on a specific hardware. 
    We first show that we can use a black-box approach to estimate a realistic latency model for a specific inference platform, without the need for low-level access to the inference computation. Then, we design a pairwise MRF to score any channel configuration and use dynamic programming to efficiently decode the best performing configuration, yielding an optimal solution for the network width search. Finally, we propose an adaptive channel configuration sampling scheme to gradually specialize the training phase to the target computational constraints. Experiments on ImageNet classification show that our approach can find networks fitting the resource constraints on different target platforms while improving accuracy over the state-of-the-art efficient networks.
\end{abstract}

\section{Introduction}

Neural networks define the state of the art in computer vision for a
wide variety of tasks. Increasingly sophisticated deep learning-based
vision algorithms are being deployed on various target platforms, but
they must be adapted to the platform-dependent latency/memory
requirements and different hardware profiles. This motivates the need
for 
task-aware neural architecture search (NAS)
methods~\cite{Angeline1994,Zoph2016,nasSurvey}.

\begin{figure}
    \centering
    \includegraphics[width=0.99\linewidth]{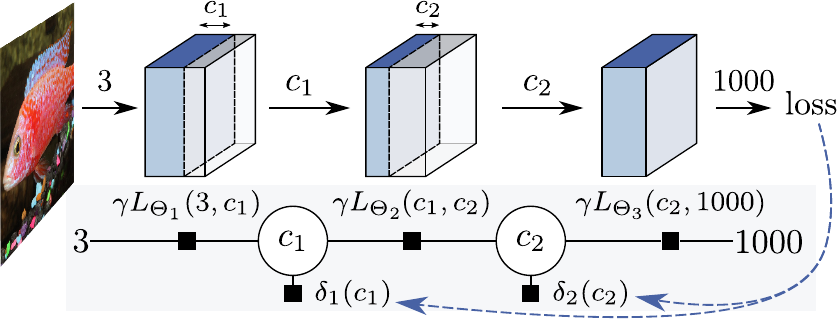}
    \caption{Overview of OWS applied to a 3-layer neural network. \emph{Top:} slimmable network; \emph{bottom:} MRF for optimal selection of channel numbers $c_1$ and $c_2$.}
    \label{fig:splash}
\end{figure}

Multiple NAS approaches have been proposed in the literature and
successfully applied to image
recognition~\cite{Cai2018ProxylessNASDN,mnasnet,Stamoulis2019SinglePathND,Liu_2018_ECCV,Wu_2019_CVPR,ZophVSL17}
and language modeling tasks~\cite{Zoph2016}. 
Despite their impressive performance, 
many of these approaches are prohibitively
expensive, requiring the training of thousands of architectures in order to find a best performing
model~\cite{Zoph2016,mnasnet,ZophVSL17,Liu_2018_ECCV,Pham2018ENAS}. 
Some methods therefore try to dramatically reduce compute overhead by
summarizing the entire search space using a single over-parametrized
neural network~\cite{Stamoulis2019SinglePathND,Wu_2019_CVPR}. 
AutoSlim~\cite{2019autoslim} nests the entire search space (varying channel numbers) in a single \emph{slimmable network} architecture~\cite{Yu2018SlimmableNN,Yu2019UniversallySN}, trained to operate at different channel number configurations at test time. 

In this work, we build on the concept of slimmable
  networks and propose a
novel adaptive optimal width search (AOWS) for efficiently searching neural network channel configurations. 
We make several key contributions. 
First, we introduce a simple black-box latency modeling method that allows to estimate a realistic latency model for a specific hardware and inference modality, without the need for low-level access to the inference
computation. 
Second, we design an optimal width search (OWS) strategy, 
using dynamic programming to efficiently decode the best performing channel configuration in a pairwise Markov random field (MRF). 
We empirically show that considering the entire channel configuration search space results into better NAS solutions compared to a greedy iterative trimming procedure~\cite{2019autoslim}. 
Third, we propose an adaptive channel configuration sampling scheme.  
This approach gradually specializes the NAS proxy to our specific target at training-time, leading to an improved accuracy-latency trade-off in practice. 
Finally, we extensively evaluate AOWS on the ImageNet classification task for 3 target platforms and show significant accuracy improvements over state-of-the-art efficient networks.

\paragraph{Related work.} The last years have seen a growing
interest for automatic neural architecture search (NAS)
methods~\cite{Angeline1994,Zoph2016,nasSurvey}. Multiple NAS
approaches have been proposed and successfully applied to image
recognition~\cite{Cai2018ProxylessNASDN,mnasnet,Stamoulis2019SinglePathND,Liu_2018_ECCV,Wu_2019_CVPR,ZophVSL17,Pham2018ENAS}
and language modeling tasks~\cite{Zoph2016}. Pioneer
approaches~\cite{Zoph2016,ZophVSL17} use reinforcement learning to
search for novel architectures with lower FLOPs and improved
accuracy. MNasNet~\cite{mnasnet} directly searches network
architecture for mobile devices. They sample a few thousand models
during architecture search, train each model for a few epochs only and
evaluate on a large validation set to quickly estimate potential model
accuracy. 
Many of these approaches require very heavy computations, 
and therefore resort to proxy tasks (e.g. small number of epochs, smaller datasets, reduced search space) before selecting the top-performing building blocks for further learning on large-scale target task~\cite{mnasnet,Pham2018ENAS,ZophVSL17}. 
To overcome these
limitations, one group of methods
directly learns the architectures for large-scale target tasks and
target hardware platforms. For instance,~\cite{Cai2018ProxylessNASDN}
assumes a network structure composed of blocks (\eg{} MNasNet~\cite{mnasnet}) and relies
on a gradient-based approach, similar to DARTS~\cite{Liu2018Darts}, to search inside each block. 
Another group of methods intends to dramatically reduce compute
overhead by summarizing the entire search space using a single
over-parametrized neural
network~\cite{Stamoulis2019SinglePathND,Wu_2019_CVPR,Yu2018SlimmableNN,Yu2019UniversallySN}.
Single-path NAS~\cite{Stamoulis2019SinglePathND} uses this principle of
nested models and combines search for channel numbers with a search
over kernel sizes. 
However, single-path NAS restricts the search over
channel numbers to 2 choices per layer
and only optimizes over a subset of the channel numbers of the network,
fixing the backbone channel numbers and optimizing only the
expansion ratios of the residual branches in architectures such as
Mobilenet-v2~\cite{Sandler2018MobileNetV2IR}. 

AutoSlim~\cite{2019autoslim} uses a \emph{slimmable network} architecture~\cite{Yu2018SlimmableNN,Yu2019UniversallySN}, 
which is trained to operate at
different channel number configurations, as a model for the
performance of a network trained to operate at a single channel
configuration. 
Thus, the entire search space (varying channel numbers)
is nested into one unique network. 
Once the slimmable network is trained, AutoSlim selects the final channel numbers with a greedy iterative trimming procedure, starting from the maximum-channel number configuration, until the resource constraints are met. 
Our approach is closely related to AutoSlim, as we also build on slimmable networks. 
In \cref{sec:autoslim}, we further detail these prior works \cite{Yu2019UniversallySN,Yu2018SlimmableNN,2019autoslim}, highlighting their similarities and differences with our approach, which we introduce in \cref{sec:latency-model,sec:viterbi-autoslim,sec:dynamic-autoslim}.

\section{Neural architecture search}\label{sec:NAS}
We now briefly outline the NAS problem statement. A general NAS problem
can be expressed as:
\begin{problem}[NAS problem]\label{pbm:nas_problem}
Given a search space $\S$, a set of resource constraints $\C$, $\text{minimize } \Delta(N)$ for $N\in\S\cap\C$. 
\end{problem}
\noindent In a supervised learning setting, the error $\Delta(N)$ is typically
defined as the error on the validation set after training network $N$
on the training set. In the following, we discuss the choices of the
search space $\S$ and of the constraint set $\C$.

\paragraph{Search space.} 
The hardness of the NAS problem depends on the search space. 
A neural network in $\S$ can be represented by its computational graph, types of each node in
the graph, and the parameters of each node.
More specialized NAS approaches fix the neural network connectivity graph and operations but aim at finding the right parameters for these operations, e.g. kernel sizes, or number of input/output channels (\emph{width}) per layer in the network.
Single-path NAS~\cite{Stamoulis2019SinglePathND} searches for kernel
sizes and channel numbers, while AutoSlim~\cite{2019autoslim} searches
for channel numbers only.  The restriction of the NAS problem to the
search of channel numbers allows for a much more fine-grained search
than in more general NAS methods.
Furthermore, channel number calibration is essential to the
performance of the network and is likely to directly affect the
inference time.

Even when searching for channel numbers only, the size of the search
space is a challenge: if a network $N$ with $n$ layers is parametrized by its channel
numbers $(c_0, \ldots, c_{n})$, where $c_i$ can take values among a
set of choices $C_i \subseteq \mathbb{N}$, the size of the design space
\begin{equation}\label{eq:search_space}
    \S_{\text{channels}} = \left\{N(c_0, c_1, \ldots, c_{n}), c_i\in C_i\right\}
\end{equation}
is exponential in the number of layers\footnote{For ease of notation, we adopt $C_0=\{I\}$ and $C_n=\{O\}$ where $I$ is the number of input channels of the network and $O$ its output channels, set by the application (\eg{} $3$ and $1000$ resp. in an ImageNet classification task)}. 
Therefore, efficient methods are needed to explore the search space, \eg{} by relying on approximations, proxies, or by representing many elements of the search space using a single network.

\paragraph{Resource constraints.}
The resource constraints $\C$ in the NAS problem
(\cref{pbm:nas_problem}) 
are hardware- and inference engine-specific constraints used in the
target application. $\C$ considered by many NAS approaches is
a bound on the number of FLOPs or performance during a single
inference. While FLOPs can be seen as a metric broadly encompassing
the desired physical limitations the inference is subjected to (\eg
latency and power consumption), it has been shown that FLOPs correlate
poorly with these end metrics~\cite{Yang2016DesigningEC}.
Therefore, specializing the NAS to a particular inference engine and
expressing the resource constraints as a bound on the target platform
limitations is of particular interest. This has given a rise to more
resource-specific NAS approaches, using resource constraints of the
form
\begin{equation}
    \C = \{N|M(N) < M_T\}
\end{equation}
where $M(N)$ is the resource metric and $M_T$ its target.  $M(N)$ can
represent latency, power consumption constraints, or combinations of
these objectives~\cite{Yang2016DesigningEC,Yang2018NetAdaptPN}. 
Given the size of the search space, NAS often requires evaluating the
resource metric on a large number of networks during the course of the
optimization. 
This makes it often impracticable to rely on 
performance measurements on-hardware during the search. Multiple methods therefore
rely on a model, such as a latency
model~\cite{Stamoulis2019SinglePathND}, which is learned beforehand
and maps a given network $N$ to an expected value of the resource
metric $M(N)$ during the on-hardware inference.

\section{Slimmable networks and AutoSlim \label{sec:autoslim}}
We now briefly review slimmable networks~\cite{Yu2018SlimmableNN} and the AutoSlim~\cite{2019autoslim}
approach.

\paragraph{Slimmable networks.}  
Slimmable neural network
training~\cite{Yu2018SlimmableNN} is designed to produce models that can
be evaluated at various network widths at test time to account for
different accuracy-latency trade-offs. At each training iteration $t$ a
random channel configuration $\vc^t = (c_0^t, \ldots, c_n^t)$ is
selected, where each channel number $c_i$ is picked among a set of
choices $C_i$ representing the desired operating channels for layer
$i$.  This allows the optimization to account for the fact that number of channels will be selected dynamically at test time. The so-called \emph{sandwich rule} (where each iteration minimizes the error of the maximum and minimum size networks in addition to a random configuration) and \emph{in-place
  distillation} (application of knowledge-distillation \cite{hinton2015distilling} between the maximum network and smaller networks) have been further introduced
by \cite{Yu2019UniversallySN} to improve slimmable network training
and increase accuracy of the resulting networks. Dynamically selecting channel numbers at
test time requires re-computing of batch normalization statistics. \cite{Yu2019UniversallySN} showed that for large batch
sizes, these statistics can be estimated using the inference of a
single batch, which is equivalent to using the batch normalization
layer in \emph{training mode} at test time.

\paragraph{Channel number search.} A slimmable network is used for the determination of the
optimized channel number configurations under specified
resources constraints.  This determination relies on the following
assumption:
\begin{assumption}[Slimmable NAS assumption]\label{asp:autoslim}
The performance of a slimmable network evaluated for a given channel
configuration $\vc\in C_0\times\ldots\times C_n$ is a good proxy for
the performance of a neural network trained in a standard fashion with
only this channel configuration.
\end{assumption}

\noindent 
Given this assumption, AutoSlim proposes a greedy iterative trimming scheme in order to select the end channel configuration from a trained slimmable network. 
The procedure starts from the maximum channel configuration $\vc=\vM$. 
At each iteration:
\begin{itemize}
    \item The performance of channel configuration $\vc^{\prime k}=(c_0, \ldots, c_{k-1}, d, c_{k+1}, \ldots, c_n)$ with $d = \max_{c < c_k} C_k$ is measured on a validation set for all $k \in [1, n-1]$;
    \item The configuration among $(\vc^{\prime k})_{k=1\ldots n-1}$ that least increases the validation error is selected for next iteration.
\end{itemize}
This trimming is repeated until the resource constraint $M(N(\vc)) < M_T$ is met. 
The output of AutoSlim is a channel configuration $\vc$ that satisfies the resource constraint, which is then trained from scratch on the training set. 

\paragraph{Discussion.} Reliance on the one-shot slimmable 
network training makes AutoSlim training very efficient, while channel
configuration inference via greedy iterative slimming is also
performed efficiently by using only one large batch per tested
configuration~\cite{2019autoslim}. The greedy optimization strategy
employed by AutoSlim is known to yield approximation guarantees with
respect to an optimal solution for resource constrained performance
maximization under certain assumptions on the underlying objective,
notably submodularity \cite{fujishige2005submodular}.  However, in
practice, optimization of a slimmable network configuration does not
satisfy submodularity or related conditions, and the employment of an
iterative greedy algorithm is heuristic.

In this work we also build on the ideas of slimmable network
training. However, in contrast to AutoSlim, we show that better NAS
solutions can be found by employing a non-greedy optimization scheme
that considers the entire channel configuration search space and
efficiently selects a single channel configuration meeting the
resource requirements. This is achieved through the use of a
Lagrangian relaxation of the NAS problem, statistic
aggregation during training, and Viterbi decoding
(\cref{sec:viterbi-autoslim}). Selecting optimal channel configuration
under available compute constraints requires precise hardware-specific
latency model. Thus in \cref{sec:latency-model} we propose an accurate
and simple black-box latency estimation approach that allows to obtain
a realistic hardware-specific latency model without the need for
low-level access to the inference computation. Finally, we propose a biased path sampling to progressively reduce the search
space at training time, allowing a gradual specialization of the
training phase to fit the target computational constraints. Our
dynamic approach (\cref{sec:dynamic-autoslim}) specializes the NAS
proxy to our specific target and leads to improved accuracy-latency
trade-offs in practice.

\section{Black-box latency model for network width search \label{sec:latency-model}}

We propose a latency model suited to the quick evaluation of the latency of a network $L(N)$ with varying channel numbers, which we use in our method. 
While other works have designed latency models~\cite{Hanhirova2018LTC,Stamoulis2019SinglePathND,VenierisBouganisFPL2017}, creating an accurate model for the fine-grained channel number choices allowed by our method is challenging. 
In theory, the FLOPs of a convolutional layer  scale as
\begin{equation}
c_{\text{in}}c_{\text{out}}WHk^2/s^2,
\end{equation}
where $c_{\text{in}}$, $c_{\text{out}}$ are input and output channel numbers, $(W, H)$ are the input spatial dimensions, $k$ is the kernel size and $s$ the stride. 
However, the dependency of the latency measured in practice to the number of FLOPs is highly non-linear. 
This can be explained by various factors: \begin{enumerate*}[label=(\roman*)] 
\item parallelization of the operations make the latency dependent on external factors, such as the number of threads fitting on a device for given parameters;
\item caching and memory allocation mechanisms are function of the input and output shapes;
\item implementation of the operators in various inference libraries such as CuDNN or TensorRT are tuned towards a particular choice of channel numbers
\end{enumerate*}.

Rather than attempting to model the low-level phenomena that govern the dependency between the channel numbers and the inference time, we use a look-up table modelling the latency of each layer in the network as a function of the channel numbers. 
For each layer $i=0\ldots n-1$, we encode as $\Theta_i$ the layer parameters that are likely to have an impact on the layer latency. 
In the case of the mobilenet-v1 network used in our experiments, we used $\Theta_i = (H, W, s, k, dw)$, where $H\times W$ the layer input size, $s$ its stride, $k$ its kernel size and $dw\in\{0, 1\}$ an indicator of the layer type: fully convolutional, or pointwise + depthwise convolutional. 
We assume that the latency can be written as a sum over layers
\begin{equation}\label{eq:lat}
    L(N(c_0, \ldots, c_n)) = \sum_{i=0}^{n-1} L_{\Theta_{i}}(c_i, c_{i+1}) ,
\end{equation}
where each layer's latency depends on the input and output channel numbers $c_i$, $c_{i+1}$ as well as the fixed parameters $\Theta_i$. 

Populating each element $L_{\Theta_{i}}(c_i, c_{j})$ in the lookup table is non-trivial. 
The goal is to measure the contribution of each individual layer to the global latency of the network. However, the measure of the inference latency of one layer in isolation includes a memory allocation and CPU communication overhead that is not necessarily present once the layer is inserted in the network. 
Indeed, memory buffers allocated on the device are often reused across different layers.

We therefore profile 
entire networks, rather than profiling individual layers in isolation. 
We measure the latency of a set of $p$ channel configurations $(\vc^1 \ldots \vc^p)$ such that each individual layer configuration in our search space
\begin{equation}\label{eq:lat_variables}
\{L_{\Theta_i}(c_i, c_{i+1}), i\in[0, n-1], c_i\in C_i, c_{i+1} \in C_j\}
\end{equation}
is sampled at least once. 
This sampling can be done uniformly among channel configurations, or biased towards unseen layer configurations using dynamic programming, as detailed in \cref{sec:latency_sampling}. 
As a result, we obtain a set of measured latencies
$
	(L(N(\vc^j)) = l_j)_{j=1\ldots P}
$, 
which by \cref{eq:lat} yield a linear system in the variables of our latency model $L_{\Theta_i}(c_i, c_{i+1})$
\begin{equation}\label{eq:linear_lat}
	\sum_{i=0}^{n-1} L_{\Theta_i}(c_i^j, c_{i+1}^j) = l_j \quad\forall j=1\ldots P.
\end{equation}
This system can be summarized as $\mA\vx = \vl$ where $\mA$ is a sparse matrix encoding the profiled configurations, $\vl$ is the corresponding vector of measured latencies and $\vx$ contains all the variables in our latency model (i.e. the individual layer latencies in \cref{eq:lat_variables}). 
We solve the linear system using  least-squares  
to obtain the desired individual layer latencies. 

We have found that this ``black-box'' approach results in a very accurate latency model for the search of channel numbers. 
The method is framework-agnostic and does not depend on the availability of low-level profilers on the inference platform. Moreover, access to a low-level profiler would still require solving the problem of assigning the memory allocation and transfers to the correct layer in the network. Our approach deals with this question automatically, and optimally assigns these overheads in order to best satisfy the assumed latency model of \cref{eq:lat}.

The solution to linear system in \cref{eq:linear_lat} can be slightly improved by adding monotonicity priors, enforcing inequalities of the form $L_{\Theta_i}(c_i, c_k) \leq L_{\Theta_i}(c_j, c_k)$ if $c_i < c_j$ and $L_{\Theta_i}(c_i, c_k) \leq L_{\Theta_i}(c_i, c_l)$ if $c_k < c_l$, 
as one expects the latency to be increasing in the number of input/output channels of the layer. 
Similar inequalities can be written between configurations with differing input sizes. 
It is straightforward to write all these inequalities as $\mV\vx \leq \vzero$ where $\mV$ is a sparse matrix, and added to the least-squares problem. 
Rather than enforcing these inequalities in a hard way, we found it best to use a soft prior, which translates into 
\begin{equation}
	\min_\vx \|\mA\vx - \vl\|^2 + \lambda \|\!\max(\mV\vx, \vzero) \|_1,
\end{equation}
where the weighting parameter $\lambda$ is set using a validation set;
this minimization can be solved efficiently using a second-order cone program solver \cite{cvxpy,ocpb:16}.

\section{Optimal width search (OWS) via Viterbi inference \label{sec:viterbi-autoslim}}
For the special case of optimizing the number of channels under a latency constraint, the NAS \cref{pbm:nas_problem} writes as
\begin{equation}
    \min_{\vc\in C_0\times\ldots\times C_n}  \Delta(N(\vc)) \quad\text{s.t. }L(N(\vc))<L_T
\end{equation}
with $L_T$ our latency target. 
We consider the following Lagrangian relaxation of the problem:
\begin{equation}\label{eq:lagrange}
    \max_\gamma \min_{\vc}  \Delta(N(\vc)) + \gamma (L(N(\vc)) - L_T)
\end{equation}
with $\gamma$ a Lagrange multiplier, similar to the formulation proposed by~\cite{srivastava19compression} for network compression. 
If the subproblems
\begin{equation}\label{eq:lagrange_subproblem}
    \min_{\vc}  \Delta(N(\vc)) + \gamma L(N(\vc))
\end{equation}
can be solved efficiently, the maximization in \cref{eq:lagrange} can be solved by binary search over $\gamma$ by using the fact that the objective is concave in $\gamma$~\cite[prop. 5.1.2]{bertsekas1995nonlinear}. This corresponds to setting the runtime penalty in \cref{eq:lagrange_subproblem} high enough that the constraint is satisfied but no higher. 

Our key idea to ensure that 
\cref{eq:lagrange_subproblem} can be solved efficiently is to find an estimate of the error of a network that decomposes over the individual channel choices as
$
    \Delta(N(\vc)) \approx \sum_{i=1}^{n-1} \delta_i(c_i);
$
indeed, given that our latency model decomposes over pairs of successive layers (\cref{eq:lat}), this form allows to write \cref{eq:lagrange_subproblem} as
\begin{equation}\label{eq:viterbi}
    \min_{\vc} \sum_{i=1}^{n-1} \delta_i(c_i) + \gamma \sum_{i=0}^{n-1} L_{\Theta_i}(c_i, c_{i+1}),
\end{equation}
which is solved efficiently by the Viterbi algorithm \cite{viterbi1967error} applied to the pairwise MRF illustrated in \cref{fig:splash}

We leverage this efficient selection algorithm in a procedure that we detail in the remainder of this section. 
As in \cref{sec:autoslim}, we train a slimmable network. 
In order to ensure faster exploration of the search space, rather than sampling one unique channel configuration per training batch, we sample a different channel configuration separately for each element in the batch. 
This can be implemented efficiently at each layer $i$ by first computing the ``max-channel'' output for all elements in the batch, before zeroing-out the channels above the sampled channel numbers for each individual element. 
This batched computation is in aggregate faster than a separate computation for each element. 

For each training example $\vx^{(t)}$, a random configuration $\vc^{(t)}$ is sampled, yielding a loss $\ell(\vx^{(t)}, \vc^{(t)})$; we also retain the value of the loss corresponding to the maximum channel configuration $\ell(\vx^{(t)}, \vM)$ -- available due to \emph{sandwich rule} training (\cref{sec:autoslim}). 
For each $i=1\ldots n-1$, we consider all training iterations $T_i(c_i) = \{t \, | \, c^{(t)}_i = c_i\} \subseteq \mathbb{N}$ where a particular channel number $c_i\in C_i$ was used.
We then define
\begin{equation}\label{eq:average_error}
    \delta_i(c_i) =  \frac{1}{|T_i(c_i)|}\sum_{t\in T_i(c_i)} \ell(\vx^{(t)}, \vc^{(t)}) - \ell(\vx^{(t)}, \vM) 
\end{equation}
as the per-channel error rates in \cref{eq:viterbi}. Measuring the loss relative to the maximum configuration loss follows the intuition that good channel numbers lead to lower losses on average. 
Empirically, we found that computing the average in \cref{eq:average_error} over the last training epoch yields good results.

\Cref{eq:viterbi} is designed for efficient inference by neglecting the interaction between the channel numbers of different layers. 
We show in our experiments (\cref{sec:experiments}) that this trade-off between inference speed and modeling accuracy compares favorably to the greedy optimization strategy described in \cref{sec:autoslim}. 
On the one hand, the number of training iterations considered in \cref{eq:average_error} is sufficient to ensure that the per-channel error rates are well estimated. Approaches that would consider higher-order interactions between channel numbers would require an exponentially higher number of iterations to achieve estimates with the same level of statistical accuracy. 
On the other hand, this decomposition allows the performance of an exhaustive search over channel configurations using the Viterbi algorithm (\cref{eq:viterbi}). 
We have observed that this selection step takes a fraction of a second and does not get stuck in local optima as occurs when using a greedy approach. The greedy approach, by contrast, took hours to complete.

\section{Adaptive refinement of Optimal Width Search (AOWS) \label{sec:dynamic-autoslim}}
We have seen in \cref{sec:viterbi-autoslim} how layerwise modeling and Viterbi inference allows for an efficient global search over configurations. 
In this section, we describe how this efficient selection procedure can be leveraged in order to refine the training of the slimmable network thereby making \cref{asp:autoslim} more likely to hold.

Our strategy for adaptive refinement of the training procedure stems from the following observation: during the training of the slimmable model, by sampling uniformly over the channel configurations we visit many of configurations that have a latency greater than our objective $L_T$, or that have a poor performance according to our current channel estimates $\delta_i(c_i)$.
As the training progresses, the sampling of the channel configurations should be concentrated around the region of interest in the NAS search space.

\begin{figure}
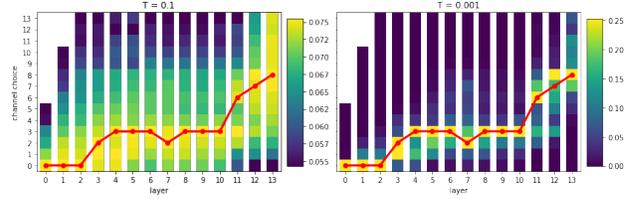

    \centering
    \includegraphics[width=0.98\linewidth]{{{figures/path_merged}}}
    \caption{Min-sum relaxation at temperatures $T=0.1$ (left) and $T=0.001$ (right). Red: min-sum path. Colored: marginal sampling probabilities. The sampled configurations approach the min-sum path as $T\rightarrow 0$.\label{fig:sampling_proba}}
\end{figure}

In order to refine the sampling around the solutions close to the minimum of \cref{eq:viterbi}, we relax the Viterbi algorithm (min-sum) using a differentiable dynamic programming procedure described in \cite{Mensch2018DifferentiableDP}. 
This strategy relaxes the minimization in \cref{eq:viterbi} into a smoothed minimization, which we compute by replacing the $\min$ operation by a $\log$-$\mathrm{sum}$-$\exp$ operation in the Viterbi forward pass. 
The messages sent from variable $c_i$ to variable $c_{i+1}$ become
\begin{equation}
\begin{aligned}
m({c_{i+1}}) = \log\sum_{c_i}\exp-\frac{1}{T}\big(&m(c_i) + \delta_i(c_{i+1}) \\[-1em]
&+ \gamma L_{\Theta_i}(c_i, c_{i+1})\big) ,
\end{aligned}
\end{equation}
where $T$ is a temperature parameter that controls the smoothness of the relaxation. 
The forward-backward pass of the relaxed min-sum algorithm yields log-marginal probabilities $\log p_i(c_i)$ for each layer whose mass is concentrated close to configurations minimizing \cref{eq:viterbi}. 
For $T=1$, these correspond to the marginal probabilities of the pairwise CRF defined by the energy of \cref{eq:viterbi}.
In the limit $T\rightarrow 0$, the probabilities become Dirac distributions corresponding to the MAP inference of the CRF as computed by the Viterbi algorithm (\cref{fig:sampling_proba}).

We introduce the following dynamic training procedure. 
First, we train a slimmable network for some warmup epochs, using uniform sampling of the configurations as in \cref{sec:autoslim}. 
We then turn to a biased sampling scheme.
We initially set $T=1$. 
At each iteration, we
\begin{enumerate}
    \item sample batch configurations according to the marginal probabilities $p_i(c_i)$,
    \item do a training step of the network,
    \item update the unary statistics (\cref{eq:average_error}),
    \item decrease $T$ according to an annealing schedule.
\end{enumerate}

This scheme progressively favours configurations that are close to minimizing \cref{eq:viterbi}. 
This reduction in diversity of the channel configurations ensures that:
\begin{itemize}
    \item training of the slimmable model comes closer to the training of a single model, thereby making \cref{asp:autoslim} more likely to hold;
    \item per-channel error rates (\cref{eq:average_error}) are averaged only over relevant configurations, thereby enforcing an implicit coupling between the channel numbers of different layers in the network.
\end{itemize}
Our experiments highlight how this joint effect leads to channel configurations with a better accuracy/latency trade-off.

\section{Experiments \label{sec:experiments}}

\begin{table}[hbt!]
\caption{Run-time vs. accuracy comparison for timings obtained with batch size 64. 
The GPU+TRT column lists the latencies and speedups allowed by TensorRT. AOWS is obtained with Mobilenet-v1/TensorRT optimization with $L_T = 0.04$ms, and a longer training schedule of $480$ epochs.\label{tbl:mobilenetv1_TRT}}
\begin{center}
\begin{adjustbox}{max width=1.0\textwidth}
\begin{tabular}{@{}lcr@{ }lc@{}}
\toprule
Method & GPU & \multicolumn{2}{c}{GPU+TRT} & Top-1   \\ 
      & ms/fr & ms/fr & $_{\text{speedup}}$ & Error (\%)  \\ \midrule
AOWS &  0.18 & \textbf{0.04}  &$_{\text{4.5x}}$& 27.5 \\
AutoSlim~\cite{2019autoslim} &  0.15 & \textbf{0.04}  &$_{\text{3.75x}}$ & 28.5 \\
Mobilenet-v1 \cite{Howard2017MobileNetsEC} &  0.25 & 0.05  &$_{\text{5x}}$ &  29.1 \\
Shufflenet-v2 \cite{Ma2018ShuffleNetVP} &  0.13 & 0.07  &$_{\text{1.9x}}$ & 30.6 \\
MNasNet~\cite{mnasnet} & 0.26 & 0.07 & $_{\text{3.7x}}$ & 26.0 \\
SinglePath-NAS~\cite{Stamoulis2019SinglePathND} & 0.28 & 0.07 & $_{\text{4.0x}}$ & 25.0 \\
ResNet-18 \cite{He2015Resnet} & 0.25  &  0.08 & $_{\text{3.1x}}$ & 30.4\\
FBNet-C~\cite{Wu_2019_CVPR}& 0.32 & 0.09 & $_{\text{3.6x}}$ & 25.1\\
Mobilenet-v2~\cite{Sandler2018MobileNetV2IR}&  0.28 & 0.10 & $_{\text{2.8x}}$ & 28.2\\
Shufflenet-v1~\cite{Zhang2017ShuffleNetAE} &  0.21 & 0.10 & $_{\text{2.1x}}$ & 32.6 \\
ProxylessNAS-G~\cite{Cai2018ProxylessNASDN} & 0.31 & 0.12 & $_{\text{2.7x}}$ & 24.9 \\
DARTS~\cite{Liu2018Darts} & 0.36 & 0.16 & $_{\text{2.3x}}$ & 26.7\\
ResNet-50 \cite{He2015Resnet} & 0.83  &   0.19 & $_{\text{4.3x}}$ & 23.9\\
Mobilenet-v3-large~\cite{Howard2019SearchingFM} & 0.30 &0.20 & $_{\text{1.5x}}$ & 24.8\\
NASNet-A*~\cite{Zoph2016} & 0.60 & - &  & 26.0 \\
EfficientNet-b0 \cite{Tan2019EfficientNetRM} & 0.59 &  0.47 & $_{\text{1.3x}}$ & 23.7\\
\bottomrule
\multicolumn{5}{l}{\footnotesize * TRT inference failed due to loops in the underlying graph}
\end{tabular}
% }
\end{adjustbox}
\end{center}
\end{table}

\paragraph{Experimental setting.}
We focus on the optimization of the channel numbers of MobileNet-v1~\cite{Howard2017MobileNetsEC}. 
The network has $14$ different layers with adjustable width. 
We consider up to $14$ channel choices for each layer $i$, equally distributed between $20\%$ and $150\%$ of the channels of the original network. 
These numbers are rounded to the nearest multiple of $8$, with a minimum of $8$ channels. 
We train AOWS and OWS models for 20 epochs with batches of size 512 and a constant learning rate 0.05. 
For the AOWS versions, after 5 warmup epochs (with uniform sampling), we decrease the temperature following a piece-wise exponential schedule detailed in \cref{sec:hyperparams}. 
We train the selected configurations with a training schedule of $200$ epochs, batch size 2048, and the training tricks described in~\cite{He2018BagOT}, including cosine annealing~\cite{Loshchilov2016SGDRSG}. 

\subsection{TensorRT latency target \label{sec:trtobj}}
We first study the optimization of MobileNet-v1 under TensorRT (TRT)\footnote{\url{https://developer.nvidia.com/tensorrt}} inference on a NVIDIA V100 GPU. 
\Cref{tbl:mobilenetv1_TRT} motivates this choice by underlining the speedup allowed by TRT inference, compared to vanilla GPU inference under the MXNet framework. 
While the acceleration makes TRT attractive for production environments, we see that it does not apply uniformly across architectures, varying between 1.3x for EfficientNet-b0 and 5x for mobilenet-v1.

\begin{figure}
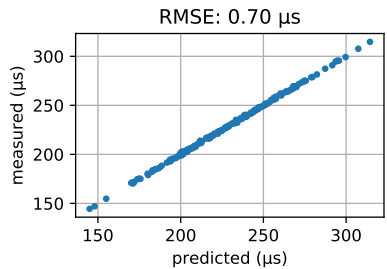

    \centering
    \includegraphics[width=0.6\linewidth]{{{figures/RMSE_TRT-crop}}}
    \caption{Measured vs. predicted latency of $200$ randomly sampled networks in our search space for the TensorRT latency model, trained using 9500 inference samples.\label{fig:latency_correlation}}
\end{figure}

\paragraph{Latency model.}
\Cref{fig:latency_correlation} visualizes the precision of our latency model as described in \cref{sec:latency-model} for $200$ randomly sampled configurations in our search space, and show that our pairwise decomposable model (\cref{sec:latency-model}) adequately predicts the inference time on the target platform. 

\begin{figure}
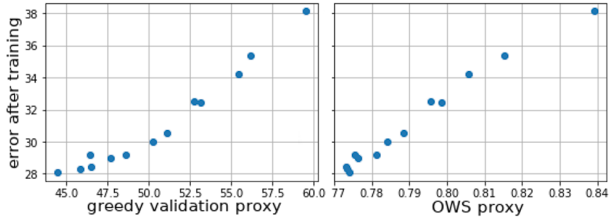

    \centering
    \includegraphics[width=0.98\linewidth]{{{figures/proxy_compare2}}}
    \caption{Comparison of the slimmable proxy used by greedy validation (left) and our error estimates used in OWS (right).
    The proxy errors of $13$ networks in our search space are compared to the final error after full training of these configurations. \label{fig:proxy_compare}}
\end{figure}

\paragraph{Proxy comparison.} 
\Cref{fig:proxy_compare} shows the correlation between the error predictor and the observed errors for several networks in our search space. 
The slimmable proxy used in AutoSlim uses the validation errors of specific configurations in the slimmable model. OWS uses the simple layerwise error model of \cref{eq:average_error}.
We see that both models have good correlation with the final error.
However, the slimmable proxy requires a greedy selection procedure, while the layerwise error model leads to an efficient and global selection.

\begin{figure}
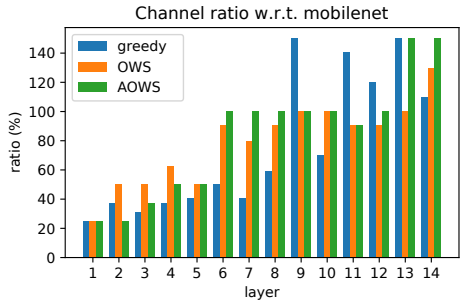

    \centering
    \includegraphics[width=0.75\linewidth]{{{figures/tristogram}}}
    \caption{Channel configurations found, as a ratio with respect to the original channels of MobileNet-v1\label{fig:channelconf_TRT}. 
    }
\end{figure}

\paragraph{Optimization results.}
We set the TRT runtime target $L_T = 0.04$ms, chosen as the reference runtime of AutoSlim mobilenet-v1. 
\Cref{tbl:final_mobilenetv1_TRT} gives the final top-1 errors obtained by the configurations selected by the different algorithms. 
\emph{greedy} reproduces AutoSlim greedy selection procedure with this TRT latency target on the slimmable proxy (\cref{sec:autoslim}). 
OWS substitutes the global selection algorithm based on channel estimates (\cref{eq:viterbi}). 
Finally, AOWS uses the adaptive path sampling procedure (\cref{sec:dynamic-autoslim}). 
\Cref{fig:channelconf_TRT} illustrates the differences between the found configurations (which are detailed in \cref{sec:foundconfs}). 
As in \cite{2019autoslim}, we observe that the configurations generally have more weights at the end of the networks, and less at the beginning, compared to the original mobilenet-v1 architecture~\cite{Howard2017MobileNetsEC}.

Despite the simplicity of the per-channel error rates, we see that OWS leads to a superior configuration over greedy, on the same slimmable model. 
This indicates that greedy selection can fall into local optimas and miss more advantageous global channel configurations. 
The AOWS approach uses the Viterbi selection but adds an adaptive refinement of the slimmable model during training, which leads to superior final accuracy. 

\Cref{tbl:mobilenetv1_TRT} compares the network found by AOWS with architectures found by other NAS approaches. 
The proposed AOWS reaches the lowest latency on-par with AutoSlim~\cite{2019autoslim}, while reducing the Top-1 image classification error by 1\%. 
This underlines the importance of the proposed platform-specific latency model, and the merits of our algorithm.

\begin{table}[ht]
\center
\caption{Accuracies and latencies of channel configurations found for TRT optimization with $L_T = 0.04$ms.\label{tbl:final_mobilenetv1_TRT}}
\begin{tabular}{@{}lcc@{}}
\toprule
Method & ms/fr & Top-1 error (\%)  \\ \midrule
greedy         & 0.04 & 29.3  \\ 
OWS & 0.04 & 28.2  \\ 
AOWS & 0.04 & 27.8  \\ 
\bottomrule
\end{tabular}
\end{table}

\paragraph{AOWS training epochs.}
One important training hyperparameter is the number of training epochs of AOWS. 
\Cref{tbl:ablation_epochs_TRT} shows that training for $10$ epochs leads to a suboptimal model; however, the results at epoch $30$ are on-par with the results at epoch $20$, which motivates our choice of picking our results at epoch $20$.

\begin{table}[ht]
\center
\caption{Effect of the number of epochs when training AOWS, for TRT optimization under $L_T = 0.04$ms.\label{tbl:ablation_epochs_TRT}}
\begin{tabular}{@{}llll@{}}
\toprule
Epochs & 10 & 20 & 30 \\ \midrule
Top-1 error (\%) & 28.1 & 27.8 & 27.9 \\ \bottomrule
\end{tabular}
\end{table}

\subsection{FLOPS, CPU and GPU targets \label{sec:flopsobj}}
We experiment further with the application of AOWS to three different target constraints. 
First, we experiment with a FLOPs objective. 
The expression of the FLOPs decomposes over pairs of successive channel numbers, and can therefore be written analytically as a special case of our latency model (\cref{sec:latency-model}). 
\Cref{tbl:flops-objective} gives the FLOPs and top-1 errors obtained after end-to-end training of the found configurations. 
We note that final accuracies obtained by AOWS are on-par or better than the reproduced AutoSlim variant (greedy).
AutoSlim~\cite{2019autoslim} lists better accuracies in the 150 and 325 MFLOPs regimes; we attribute this to different choice of search space (channel choices) and training hyperparameters, which were not made public; one other factor is the use of a 480 epochs training schedule, while we limit to 200 here.

\begin{table}[t]
\caption{Optimizing for FLOPs \label{tbl:flops-objective}}
\centering
\begin{tabular}{@{}lcc@{}}
\toprule
Variant & MFLOPs & Top-1 error (\%) \\ \midrule
AutoSlim ~\cite{2019autoslim}& 150 & 32.1 \\
greedy$_{150}$ & 150 &  35.8\\
AOWS & 150 & 35.9 \\
\hline
AutoSlim~\cite{2019autoslim} & 325 & 28.5 \\
greedy$_{325}$ &  325 & 31.0 \\
AOWS& 325 & 29.7 \\
\hline
AutoSlim~\cite{2019autoslim} & 572 & 27.0 \\
greedy$_{572}$ & 572 & 27.6 \\
AOWS& 572 & 26.7 \\ \bottomrule
\end{tabular}
\end{table}

\begin{figure}[t!]
    \centering
    \begin{subfigure}[t]{0.5\columnwidth}
        \centering
        \includegraphics[width=.97\linewidth]{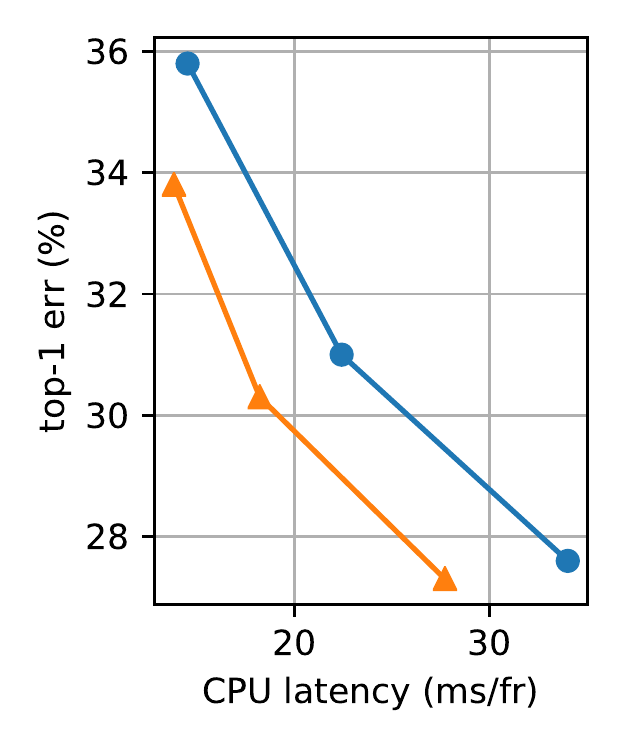}
        \caption{CPU}
    \end{subfigure}%
    ~ %
    \begin{subfigure}[t]{0.5\columnwidth}
        \centering
        \includegraphics[width=.97\linewidth]{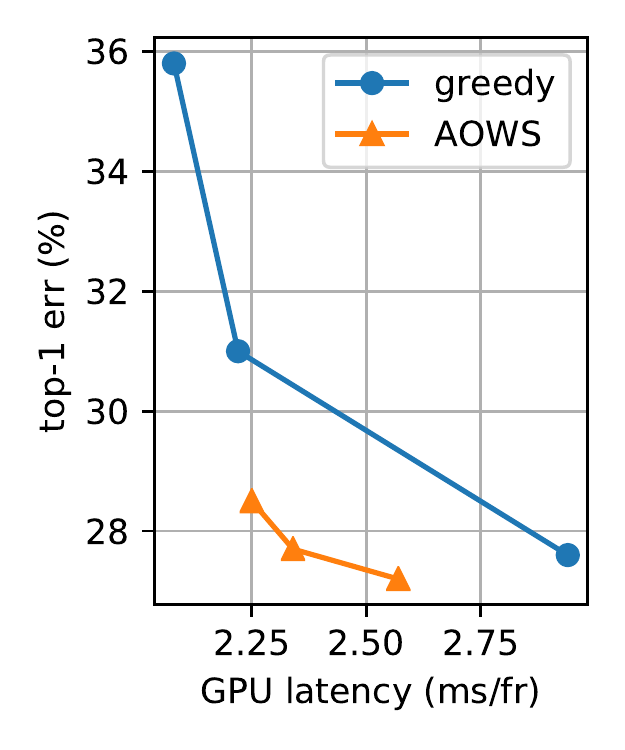}
        \caption{GPU}
    \end{subfigure}
    \caption{Pareto front of greedy, vs. Pareto front of AOWS optimized for CPU and GPU latency models.\label{fig:pareto}}
\end{figure}

We turn to realistic latency constraints, considering CPU inference on an Intel Xeon CPU with batches of size 1, and GPU inference on an NVIDIA V100 GPU with batches of size 16, under PyTorch~\cite{paszke2017automatic}.\footnote{See \cref{sec:cpugpu} for details on framework version and harware.}
\Cref{tbl:cpu-objective,tbl:gpu-objective} show the results for 3 latency targets, and the resulting measured latency. 
We also report the latencies of the channel numbers on the greedy solution space corresponding to the three configurations in \cref{tbl:flops-objective}. 
By comparison of the accuracy/latency tradeoff curves in \cref{fig:pareto}, it is clear that using AOWS leads to more optimal solutions than greedy; in general, we consistently find models that are faster and more accurate. 

We observe that the gains of AOWS over greedy are more consistent than in the case of the FLOPs optimization (\cref{sec:latency-model}). 
We note that the analytical FLOPs objective varies more regularly in the channel configurations, and therefore presents less local optima, than empirical latencies measured on-device. This might explain why the greedy approach succeeds at finding appropriate configurations in the case of the FLOPs model better than in the case of realistic latency models.

\begin{table}[t]
\caption{Optimizing for CPU latency (@ indicates the latency targets) \label{tbl:cpu-objective}}
\centering
\begin{tabular}{@{}lcc@{}}
\toprule
Variant & ms/fr & Top-1 error (\%) \\ \midrule
AOWS @ 15ms & 13.8 & 33.8 \\
AOWS @ 20ms & 18.2 & 30.3 \\
AOWS @ $30$ms & 27.7 & 27.3 \\ 
greedy$_{150}$ & 14.5 &  35.8\\
greedy$_{325}$ &  22.4 & 31.0 \\
greedy$_{572}$ & 34.0 & 27.6 \\
\bottomrule
\end{tabular}
\end{table}

\begin{table}[t]
\caption{Optimizing for GPU latency (@ indicates the latency targets) \label{tbl:gpu-objective}}
\centering
\begin{tabular}{@{}lcc@{}}
\toprule
Variant & ms/fr & Top-1 error (\%) \\ \midrule
AOWS @ 2.2ms & 2.25 & 28.5 \\
AOWS @ 2.4ms & 2.34 & 27.7 \\
AOWS @ 2.6ms & 2.57 & 27.2 \\
greedy$_{150}$ & 2.08 &  35.8\\
greedy$_{325}$ & 2.22 & 31.0 \\
greedy$_{572}$ & 2.94 & 27.6 \\
\bottomrule
\end{tabular}
\end{table}

\section{Conclusion} 
Efficiently searching for novel network architectures while optimizing
accuracy under latency constraints on a target platform and task is of
high interest for the computer vision community. In this paper we propose
a novel efficient one-shot NAS approach to optimally search CNN
channel numbers, given latency constraints on a specific hardware. To
this end, we first design a simple but effective black-box latency
estimation approach to obtain precise latency model for a specific
hardware and inference modality, without the need for low-level access
to the inference computation. Then, we introduce a pairwise MRF
framework to score any network channel configuration and use the Viterbi
algorithm to efficiently search for the most optimal solution in the
exponential space of possible channel configurations. Finally, we
propose an adaptive channel configuration sampling strategy to
progressively steer the training towards finding novel configurations that
fit the target computational constraints. Experiments on ImageNet
classification task demonstrate that our approach can find networks
fitting the resource constraints on different target platforms while
improving accuracy over the state-of-the-art efficient networks. The code has been released at \url{http://github.com/bermanmaxim/AOWS}.

\paragraph{Acknowledgements.} We thank Kellen Sunderland and Haohuan Wang
 for help with setting up and benchmarking TensorRT inference, and Jayan Eledath for useful discussions. M. Berman and M. B. Blaschko acknowledge support from the Research Foundation~-~Flanders (FWO) through project numbers G0A2716N and G0A1319N, and funding from the Flemish Government under the Onderzoeksprogramma Artifici\"{e}le Intelligentie (AI) Vlaanderen programme. 

\FloatBarrier

{\small
\bibliographystyle{ieee_fullname}
\bibliography{biblio}
}

\clearpage
\appendix
\counterwithin{figure}{section}
\counterwithin{table}{section}
\counterwithin{equation}{section}
\pagenumbering{Roman}
\pagestyle{plain}

\twocolumn[\centering
{\vspace*{0.5cm}
\Large \bf AOWS: Adaptive and optimal network width search with latency constraints\\[1.2em]
Supplementary\\[1.2em]
}\vspace{3em}]

\section{Latency model: biased sampling \label{sec:latency_sampling}}
We describe the biased sampling strategy for the latency model, as described in \cref{sec:latency-model}. 
Using the notations of \cref{sec:latency-model}, the latency model is the least-square solution of a linear system $\mA\vx = \vl$. 
The variables of the system are the individual latency of every layer configuration $L_{\Theta_i}(c_i, c_{i+1})$. 
To ensure that the system is complete, each of these layer configurations must be present at least once among the model configurations benchmarked in order to establish the latency model. 
Instead of relying on uniform sampling of the channel configurations, we can bias the sampling in order to ensure that the variable of the latency model $L_{\Theta_i}(c_i, c_{i+1})$ that has been sampled the least amount of time is present.

As in AOWS, we rely on a Viterbi algorithm in order to determine the next channel configuration to be benchmarked. 
Let $N(c_i, c_{i + 1})$ be the number of times variable $L_{\Theta_i}(c_i, c_{i+1})$ has already been seen in the benchmarked configurations, and
\begin{equation}
    M = \min_{i\in[0, n-1]} \min_{\substack{c_i \in C_i\\ c_{i+1} \in C_{i + 1}}} N(c_i, c_{i + 1})
\end{equation}
the minimum value taken by $N$.
The channel configuration we choose for the next benchmarking is the solution minimizing the pairwise decomposable energy
\begin{equation}
    \min_{c_0, \ldots, c_n} \sum_{i=0}^{n-1} -[N(c_i, c_{i+1}) = M].
\end{equation}
using the Iverson bracket notation.
This energy ensures that at least one of the least sampled layer configurations is present in the sampled configuration. 

This procedure allows to set a lower bound on the count of all variables among the benchmarked configurations. 
The sampling can be stopped when the latency model has reached an adequate validation accuracy.

\section{Optimization hyperparameters}\label{sec:hyperparams}
For the temperature parameter, we used a piece-wise exponential decay schedule, with values $1$ at epoch $5$, to $10^{-2}$ at epoch $6$, $10^{-3}$ at epoch $10$, and $5 \cdot 10^{-4}$ at epoch $20$.

\section{Framework versions and CPU/GPU models}\label{sec:cpugpu}
We detail the frameworks and hardware used in the experiments of \cref{sec:trtobj,sec:flopsobj}. 
Although we report latencies in terms of ms/frame, the latency models are estimated with batches of size bigger than 1. 
In general, we want to stick to realistic operating settings: GPUs are more efficient for bigger batches, and the batch choice impacts the latency/throughput tradeoff.

The \emph{TRT experiments} are done on an NVIDIA V100 GPU with TensorRT 5.1.5 driven by MXNet v1.5, CUDA 10.1, CUDNN 7.6, with batches of size 64. The \emph{CPU inference experiments} are done on an Intel Xeon® Platinum 8175 with batches of size 1, under PyTorch 1.3.0. The \emph{GPU inference experiments} are done on an NVIDIA V100 GPU with batches of size 16, under PyTorch 1.3.0 and CUDA 10.1.

\section{Layer channel numbers and final configuration numbers found}\label{sec:foundconfs}
In \cref{tbl:searchspace}, we detail the search space in the channel numbers described in \cref{sec:experiments}.

\begin{table}[b]
\vspace{3em}
\caption{Search space: channel configurations for all 14 layers in MobileNet-v1. The first layer always has an input with $3$ channels; the last layer always outputs $1000$ channels for ImageNet classification. 
The bold values indicate the initial MobileNet-v1 configuration numbers.  \label{tbl:searchspace}}
\centering
\begin{adjustbox}{max width=1.0\textwidth}
\begin{tabular}{@{}ll@{}}
\toprule
$i$ & $C_i$ \\ \midrule
1 & 8, 16, 24, \textbf{32}, 40, 48 \\
2 & 16, 24, 32, 40, 48, 56, \textbf{64}, 72, 80, 88, 96 \\
3 & 24, 40, 48, 64, 80, 88, 104, 112, \textbf{128}, 144, 152, 168, 176, 192 \\
4 & 24, 40, 48, 64, 80, 88, 104, 112, \textbf{128}, 144, 152, 168, 176, 192 \\
5 & 48, 80, 104, 128, 152, 176, 208, 232, \textbf{256}, 280, 304, 336, 360, 384 \\
6 & 48, 80, 104, 128, 152, 176, 208, 232, \textbf{256}, 280, 304, 336, 360, 384 \\
7 & 104, 152, 208, 256, 304, 360, 408, 464, \textbf{512}, 560, 616, 664, 720, 768 \\
8 & 104, 152, 208, 256, 304, 360, 408, 464, \textbf{512}, 560, 616, 664, 720, 768 \\
9 & 104, 152, 208, 256, 304, 360, 408, 464, \textbf{512}, 560, 616, 664, 720, 768 \\
10 & 104, 152, 208, 256, 304, 360, 408, 464, \textbf{512}, 560, 616, 664, 720, 768 \\
11 & 104, 152, 208, 256, 304, 360, 408, 464, \textbf{512}, 560, 616, 664, 720, 768 \\
12 & 104, 152, 208, 256, 304, 360, 408, 464, \textbf{512}, 560, 616, 664, 720, 768 \\
13 & 208, 304, 408, 512, 616, 720, 816, 920, \textbf{1024}, 1128, 1232, 1328, 1432, 1536 \\
14 & 208, 304, 408, 512, 616, 720, 816, 920, \textbf{1024}, 1128, 1232, 1328, 1432, 1536 \\ \bottomrule
\end{tabular}
\end{adjustbox}
\end{table}

\begin{table*}[b]
\vspace{3em}
\caption{Channel configurations found in the TRT optimization (\cref{sec:trtobj}), visualized in \cref{fig:channelconf_TRT}, and with top-1 errors given in \cref{tbl:final_mobilenetv1_TRT} in the paper. Results are compared to the original Mobilenet-v1~\cite{Howard2017MobileNetsEC} channels.
\label{tbl:solutions}}
\centering
\begin{tabular}{@{}ll@{}}
\toprule
method & configuration \\ \midrule
greedy & 8, 24, 40, 48, 104, 128, 208, 304, 768, 360, 720, 616, 1536, 1128 \\
OWS & 8, 32, 64, 80, 128, 232, 408, 464, 512, 512, 464, 464, 1024, 1328 \\
AOWS & 8, 16, 48, 64, 128, 256, 512, 512, 512, 512, 464, 512, 1536, 1536 \\ \midrule
Mobilenet-v1~\cite{Howard2017MobileNetsEC} & 32, 64, 128, 128, 256, 256, 512, 512, 512, 512, 512, 512, 1024, 1024 \\
\bottomrule
\end{tabular}
\end{table*}

\end{document}